\documentclass[a4paper,conference]{IEEEtran}  
\ifCLASSINFOpdf
\else
\fi

\usepackage{cite}
\usepackage{amsmath,amssymb,amsfonts}
\usepackage{algorithmic}
\usepackage{graphicx}
\usepackage{subcaption}
\usepackage{adjustbox}      
\usepackage{xcolor}
\usepackage{tikz}
\usepackage{float}
\usepackage{verbatim}
\usepackage{pgfplots}
\pgfplotsset{compat=1.8}
\usepackage{pgfplotstable}
\usepackage{sfmath}
\usepackage{mathdots}
\usepackage{yhmath}
\usepackage{cancel}
\usepackage{color}
\usepackage{gensymb}
\usetikzlibrary{fadings}
\usepackage{multirow}
\usepackage[normalem]{ulem}
\useunder{\uline}{\ul}{}

\DeclareMathOperator{\arctantwo}{arctan2}

\definecolor{unlabeled}{rgb}{0.0392,    0.0392,    0.0392}
\definecolor{sky}{rgb}{0.5451,    0.6471,    0.5137}
\definecolor{water}{rgb}{0.3235,    0.5980,    0.73536}
\definecolor{windows}{rgb}{0.7255,    0.7137,    0.4471}
\definecolor{road}{rgb}{0.7216,    0.5804,    0.3412}
\definecolor{car}{rgb}{0.6412,    0.3431,    0.3039}
\definecolor{buildings}{rgb}{0.2980,    0.2980,    0.2980}
\definecolor{none}{rgb}{0.7,0.7,0.7}
  
\title{
Polarimetric image augmentation
}

\author{Marc Blanchon$^{1\dag}$, Olivier Morel$^{1}$, Fabrice Meriaudeau$^{1}$, Ralph Seulin$^{1}$ and D\'esir\'e Sidib\'e$^{2}$ \\
$^{1}$ERL VIBOT CNRS 6000, ImViA, Universit\'e Bourgogne Franche-Comt\'e, 71710, Le Creusot, France\\
$^{2}$IBISC, Univ Evry, Universit\'e Paris-Saclay, 91025, Evry, France\\
$^{\dag}$Email: marc.blanchon@u-bourgogne.fr\\
}

\begin{document}

\newcommand{\fabrice}[1]{\textcolor{red}{[fabrice: #1]}} 
\newcommand{\desire}[1]{\textcolor{blue}{[desire: #1]}} 
\newcommand{\olivier}[1]{\textcolor{magenta}{[olivier: #1]}} 
\newcommand{\ralph}[1]{\textcolor{green}{[ralph: #1]}} 
\newcommand{\thibault}[1]{\textcolor{gray}{[thibault: #1]}} 
\newcommand{\yifei}[1]{\textcolor{violet}{[yifei: #1]}} 

\maketitle
\thispagestyle{empty}
\pagestyle{empty}

\begin{abstract}
Robotics applications in urban environments are subject to obstacles that exhibit specular reflections hampering autonomous navigation. On the other hand, these reflections are highly polarized and this extra information can successfully be used to segment the specular areas. In nature, polarized light is obtained by reflection or scattering. Deep Convolutional Neural Networks (DCNNs) have shown excellent segmentation results, but require a significant amount of data to achieve best performances. The lack of data is usually overcomed by using augmentation methods. However, unlike RGB images, polarization images are not only scalar (intensity) images and standard augmentation techniques cannot be applied straightforwardly. We propose to enhance deep learning models through a regularized augmentation procedure applied to polarimetric data in order to characterize scenes more effectively under challenging conditions. We subsequently observe an average of 18.1\% improvement in IoU between non augmented and regularized training procedures on real world data.
\end{abstract}


\section{Introduction}
Navigation in urban environments can be prone to errors due to highly reflective areas while using RGB cameras. On the other hand, polarization imaging can cope with such environments. Accurate navigation requires dense depth map and relies on pixel-based information versus object detection methods like bounding boxes\cite{redmon2017yolo9000} . To achieve accurate segmentation using polarimetric images as input, one requires a good and large data-set illustrating hazardous areas such as cars, water spillage, windows, etc. Unfortunately, very few polarimetric datasets are available or they do not consider scenes from urban areas.
Datasets represent a critical factor affecting learning methods and particularly deep learning ones. This need is sufficiently expressed by studying the size of different widely used datasets: ImageNet\cite{imagenet_cvpr09} contains 14 million images, CityScapes\cite{Cordts2016Cityscapes} 25,000 images, \mbox{MNIST\cite{lecun-mnisthandwrittendigit-2010}} 60,000 images etc.\\
\indent Analyzing these data, one can deduce that the vast majority of frequently used datasets are provided with data acquired in a conventional mode (RGB, grayscale, etc.). These datasets, in addition to being significant, are therefore easily transformable and expandable using standard augmentation methods (rotation, symmetry, distortion, etc.). Such methods not only increase the amount of data, they also make learning more generic and therefore avoids overfitting\cite{DBLP:journals/corr/abs-1712-04621}.
However, unlike RGB images, polarization images are not only scalar (intensity) images and standard augmentation techniques cannot be applied straightforwardly.\\
\indent In this paper, we consider the problem of artificially augmenting polarization images that are in direct relation to the physics of the scene acquired by the camera. This acquisition/scene relationship drastically complicates the expansion of a dataset with standard techniques. However, it is necessary to address this issue, as the available data on polarimetric modality is extremely limited despite its increasingly popular uses \cite{berger2017depth,cui2017polarimetric,rastgoo2018attitude,nguyen20173d}.\\
\begin{figure}[!t]
\centering
        \begin{subfigure}[b]{0.3\linewidth}
            \centering
            \includegraphics[width=\linewidth]{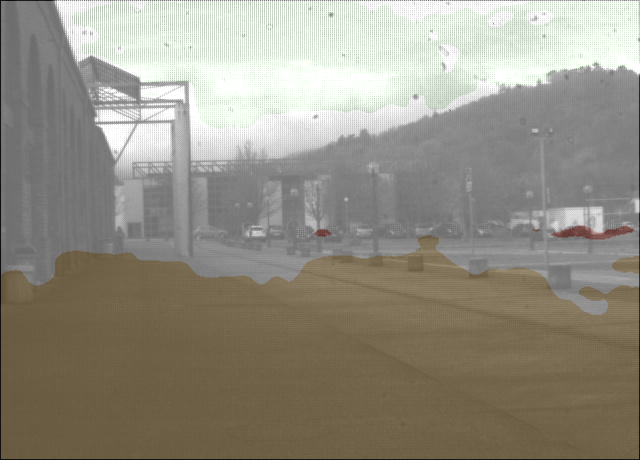}
        \end{subfigure}
        \begin{subfigure}[b]{0.3\linewidth}  
            \centering 
            \includegraphics[width=\linewidth]{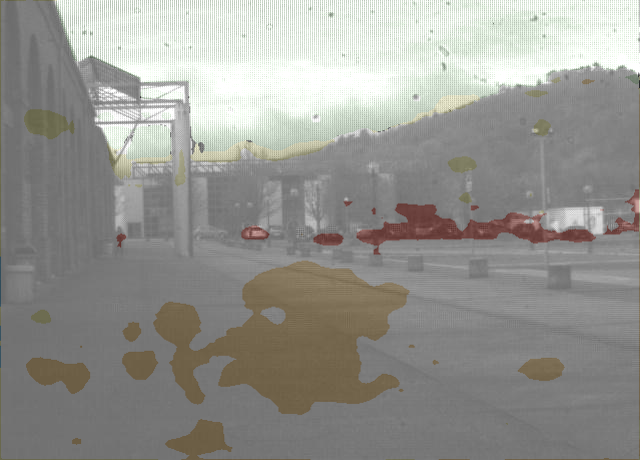}
        \end{subfigure}
        \begin{subfigure}[b]{0.3\linewidth}  
            \centering 
            \includegraphics[width=\linewidth]{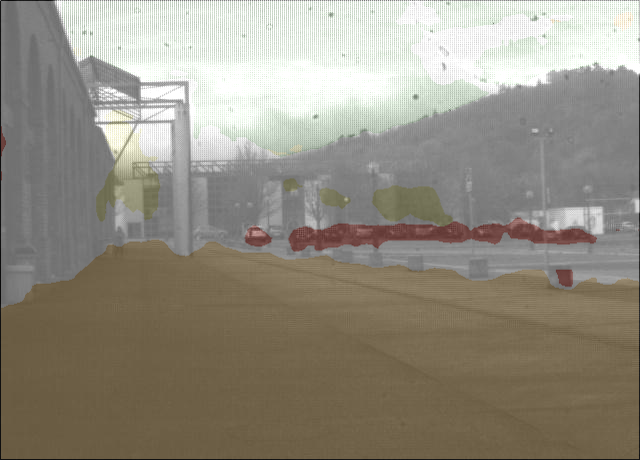}
        \end{subfigure}
        \vskip\baselineskip
        \begin{subfigure}[b]{0.3\linewidth}  
            \centering 
            \includegraphics[width=\linewidth]{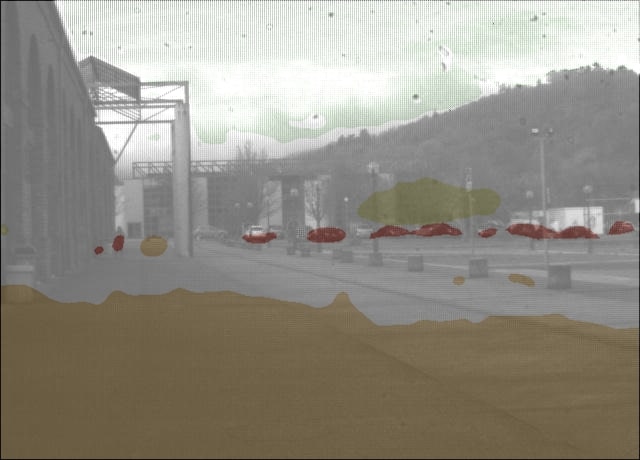}
        \end{subfigure}
        \begin{subfigure}[b]{0.3\linewidth}  
            \centering 
            \includegraphics[width=\linewidth]{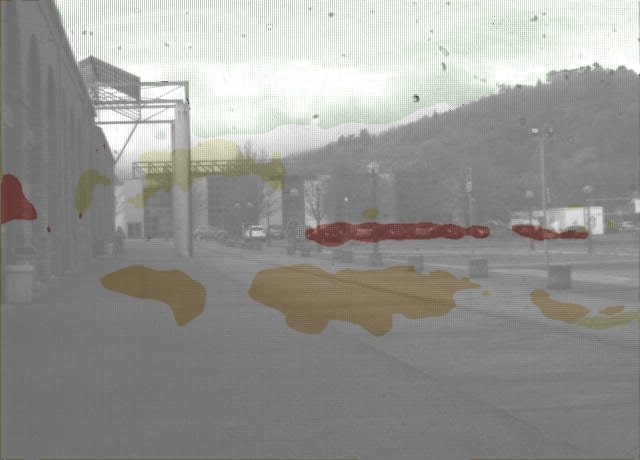}
        \end{subfigure}
        \begin{subfigure}[b]{0.3\linewidth}  
            \centering 
            \includegraphics[width=\linewidth]{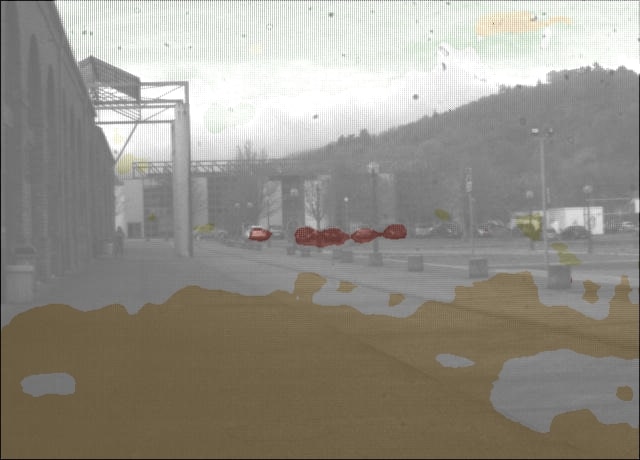}
        \end{subfigure}
\caption{Illustration of the diverse segmentation results obtained with different augmentation methods. The top line shows results for DeepLabV3 network not pre-trained, and the bottom line the results with a pre-trained network. From left to right are presented predictions from networks trained with: un-augmented dataset, standarly augmented dataset and augmented dataset following our procedure. The four most representative segmented classes here are: \textit{Road} (\textcolor{road}{dark yellow/orange}), \textit{Cars} (\textcolor{car}{red}), \textit{Sky} (\textcolor{sky}{light green}) and \textit{None} (\textcolor{none}{light grey}). 
}
\label{comp1}
\vspace{-0.25cm}
\end{figure}

\indent We have explored augmentation operations applicable to polarimetry under any condition. Initially, using the interpolation proposed by Ratliff et al.\cite{ratliff2009interpolation} combined with the channel organization proposed by Wolff and Andreou\cite{wolff1995polarization}, we propose an approach for applying rotation and/or symmetry to polarimetric images. In our case, for the experimental results, multiple trainings were performed with either a raw dataset (limited number of images), a standardly increased dataset (significant number of images without respecting physical properties) or an augmented dataset following our procedure (significant number of images with unaltered physical properties). To focus exclusively on the impact of the data, all trainings were performed using the DeepLabV3+ network\cite{chen2017rethinking}. Furthermore, we demonstrate the integrity of physical properties and the effectiveness of such a process on images that should be used for training.\\
\indent As shown in Fig. \ref{comp1}, a significant difference is then visible when training is performed successively with differently augmented datasets: raw, augmented by omitting physical properties and with a dataset using our augmentation method. 
Visually, it is possible to see a segmentation capability more adequate for the recognition of areas of interest for robotic applications. 

The paper is organized as follows.
Section \ref{sec:related} provides a brief overview of related works including pixel-wise segmentation and polarimetric image analysis.
Then, in Section \ref{sec:methods} we describe the proposed augmentation method in details.
Section \ref{exp} presents experimental results showing the effectiveness of our approach, and the paper ends with concluding remarks in Section \ref{sec:conclusion}.

\section{Related Works}\label{sec:related}
\subsection{Pixel-Wise Semantic Segmentation}
Most of the research on pixel-wise semantic segmentation use conventional imaging as input data, either RGB or depth. We can observe a constant evolution of networks accuracy due on the one side to the improvement of networks and on the other side to the increase in datasets sizes. Also, some tasks are recurrent in the community: segmentation of urban scenes\cite{DBLP:journals/corr/abs-1712-04621,paszke2016enet,lin2017refinenet,chen2017rethinking,chen2018deeplab,li2018traffic,liu2018deep}, indoor scenes understanding\cite{silberman2012indoor,cimpoi2015deep,Handa_2016_CVPR,qi2017pointnet,dai2017scannet} or medical images analysis\cite{ronneberger2015u,havaei2017brain,litjens2017survey}. The task addressed in this paper shares a common aspect with medical imaging. Indeed, one of the common disadvantages of medical images and urban scene acquired with polarimetric sensor is the lack of large annotated datasets for training.
In particular,  the specificity of the polarimetric information makes it rich but also rare.\\
Semantic segmentation being a dense and valuable information, capable of characterizing a scene at any point, it can be exploited very extensively in robotics for navigation and detection.

\subsection{Polarimetric Modality}
Polarimetry has unique properties since it allows the acquisition of changes in the state of light\cite{wolff1995polarization}. It particularly characterizes reflective areas. The attractiveness of these images is their ability to describe both the diffuse and specular parts of a scene. Considering all these advantages, polarization imaging offers a comprehensive range of possibilities that could enhance more standard modality such as RGB.\\
\indent As polarization images not exclusively carry intensity information but also information about the surface it is reflected upon, these extra informations makes it possible to extend the range of applications in computer vision, like Shape from Polarimetry, which reconstructs specular (or partially specular) objects\cite{rahmann2001reconstruction,morel2006active,morel2005polarization,cui2017polarimetric}. Other works have been carried out to improve certain approaches, such as attitude estimation\cite{shabayek2012vision,rastgoo2018attitude}, water hazard detection\cite{nguyen20173d,iqbal2009survey}, catadioptric camera calibration\cite{morel2007catadioptric}, and depth estimation\cite{kadambi2015polarized,berger2017depth}.\\
\indent Considering applications in urban environments, inferring extra knowledge about objects from the reflected light, would result in better perception and therefore useful for unmanned vehicles and robots.

\section{Method}\label{sec:methods}
\subsection{Reminder on polarimetric image processing}
\begin{figure}[!t]
\centering
\vspace{0.3cm}
   \includegraphics[width=.5\linewidth]{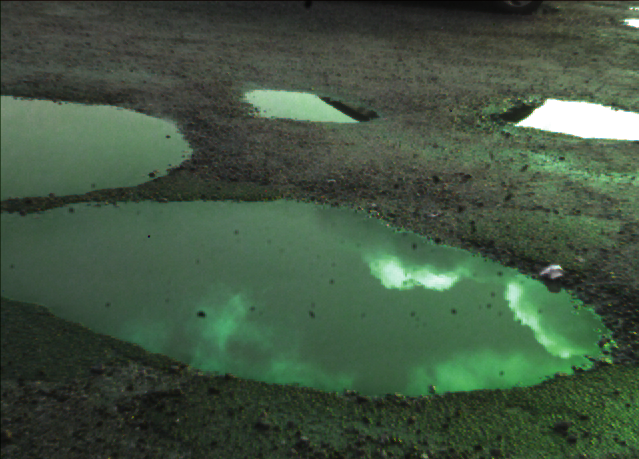}
   \caption{Illustration of the behaviour of HSL images when the scene has reflective areas. Here, four clearly visible puddles are strongly coloured green (green represents the angle of polarization, corresponding approximately to 90 degrees according to the HSL color wheel\cite{joblove1978color}).}\label{HSL}
   \vspace{-0.25cm}
\end{figure}
The acquisition is made by rotating a polarizer in front of the camera, or via a sensor exploiting the division of focal plane (DoFP). In both cases, pre-processing is frequently necessary to acquire the informative part of the images. Indeed, raw images are grayscale images; they provide very low level description of the scenes. Commonly, many approaches require a transformation either to have specific information such as angle and degree of polarization or to interpolate the images. Ratliff et al.\cite{ratliff2009interpolation} proposed a widely used approach to merge these two information. Four images acquired with four distinct polarizer angles (0$^\circ$, 45$^\circ$, 90$^\circ$, 135$^\circ$) are required for this pre-processing step, via four unique acquisitions or using a DoFP sensor. From these 4 images, $P_0$, $P_{45}$, $P_{90}$, $P_{135}$, one can reconstruct three informative images: the intensity I, the angle of polarization AoP and the degree of polarization DoP.  The last two images correspond, from a physical point of view, to the angle and power of reflection, respectively.
\begin{equation}\label{eq:DoPcalc}
    \textrm{DoP} = \frac{\sqrt{(P_0 - P_{90})^2 + (P_{45} - P_{135})^2}}{(P_0 + P_{90})},
\end{equation}
\begin{equation}\label{eq:AoPcalc}
    \textrm{AoP} = \frac{1}{2} \arctantwo \Bigg( \big(P_{45} - P_{135}\big), \big(P_0 - P_{90}\big) \Bigg), 
\end{equation}
\begin{equation}\label{eq:I}
    I = \frac{P_0 + P_{45} + P_{90} + P_{135}}{2}.
\end{equation}
These informative images are combined as proposed by Wolff and Andreou\cite{wolff1995polarization}  for display:
\begin{equation}\label{eq:aop}
    H \longrightarrow 2*\textrm{AoP},  \quad S \longrightarrow \textrm{DoP}, \quad L \longrightarrow I/255 .
\end{equation}
$H$, $S$ and $L$ are the three channels of a standard HSL colorspace (Hue, Saturation, Luminance).
As shown in Fig.\ref{HSL}, the singular characteristic of these new images is that the more reflective the surface is the more colored it appears.

\subsection{Applying transformations to polarimetric images}
Once the images have been transformed it is possible to apply augmentation processes. However, care must be taken to consider the physical properties induced by the scene (and also by the sensor).
\begin{figure}[t]
\vspace{0.3cm}
\centering
\begin{subfigure}[b]{0.32\linewidth}
            \centering
                        \caption*{Base Image}

            \includegraphics[width=\linewidth]{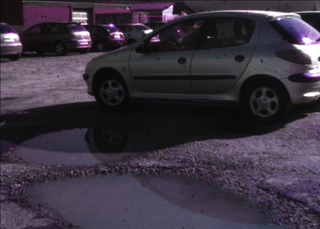}
        \end{subfigure}
        \begin{subfigure}[b]{0.32\linewidth}  
            \centering 
                        \caption*{Rotated Image}

            \includegraphics[width=\linewidth]{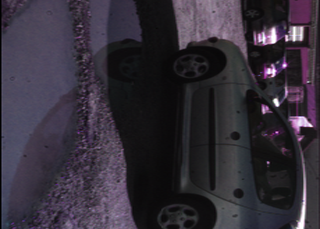}
        \end{subfigure}
        \begin{subfigure}[b]{0.32\linewidth}
            \centering
                        \caption*{Regularized Image}

            \includegraphics[width=\linewidth]{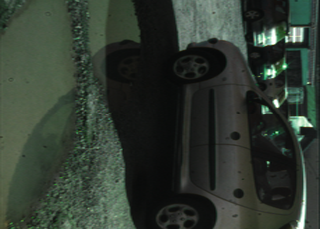}
        \end{subfigure}\\\vspace{0.3cm}\rule[1ex]{\linewidth}{0.5pt}\\\vspace{-0.3cm}
        \usetikzlibrary{patterns} 


%
%

\tikzset{
    hatch distance/.store in=\hatchdistance,
    hatch distance=10pt,
    hatch thickness/.store in=\hatchthickness,
    hatch thickness=0.3pt
}

\makeatletter
\pgfdeclarepatternformonly[\hatchdistance,\hatchthickness]{northeast}
{\pgfqpoint{0pt}{0pt}}
{\pgfqpoint{\hatchdistance}{\hatchdistance}}
{\pgfpoint{\hatchdistance-1pt}{\hatchdistance-1pt}}%
{
    \pgfsetcolor{\tikz@pattern@color}
    \pgfsetlinewidth{\hatchthickness}
    \pgfpathmoveto{\pgfqpoint{0pt}{0pt}}
    \pgfpathlineto{\pgfqpoint{\hatchdistance}{\hatchdistance}}
    \pgfusepath{stroke}
}

\pgfdeclarepatternformonly[\hatchdistance,\hatchthickness]{northwest}
{\pgfqpoint{0pt}{0pt}}
{\pgfqpoint{\hatchdistance}{\hatchdistance}}
{\pgfpoint{\hatchdistance-1pt}{\hatchdistance-1pt}}%
{
    \pgfsetcolor{\tikz@pattern@color}
    \pgfsetlinewidth{\hatchthickness}
    \pgfpathmoveto{\pgfqpoint{\hatchdistance}{0pt}}
    \pgfpathlineto{\pgfqpoint{0pt}{\hatchdistance}}
    \pgfusepath{stroke}
}

\pgfdeclarepatternformonly[\hatchdistance,\hatchthickness]{horizontal}
{\pgfqpoint{0pt}{0pt}}
{\pgfqpoint{\hatchdistance}{\hatchdistance}}
{\pgfpoint{\hatchdistance-1pt}{\hatchdistance-1pt}}%
{
    \pgfsetcolor{\tikz@pattern@color}
    \pgfsetlinewidth{\hatchthickness}
    \pgfpathmoveto{\pgfqpoint{0pt}{0pt}}
    \pgfpathlineto{\pgfqpoint{\hatchdistance}{0pt}}
    \pgfusepath{stroke}
}

\pgfdeclarepatternformonly[\hatchdistance,\hatchthickness]{vertical}
{\pgfqpoint{0pt}{0pt}}
{\pgfqpoint{\hatchdistance}{\hatchdistance}}
{\pgfpoint{\hatchdistance-1pt}{\hatchdistance-1pt}}%
{
    \pgfsetcolor{\tikz@pattern@color}
    \pgfsetlinewidth{\hatchthickness}
    \pgfpathmoveto{\pgfqpoint{0pt}{0pt}}
    \pgfpathlineto{\pgfqpoint{0pt}{\hatchdistance}}
    \pgfusepath{stroke}
}
\makeatother

\begin{center}
\resizebox{\linewidth}{!}{
\begin{tikzpicture}

\draw[pattern=horizontal,pattern color=gray,hatch distance=8pt, hatch thickness = 0.8pt]  (0,1.5) rectangle (1.5,3)node[midway]{$0^\circ$}; 
 \draw[pattern=northwest,pattern color=gray,hatch distance=10pt, hatch thickness = 0.8pt] (1.5,1.5) rectangle (3,3) node[midway]{$45^\circ$};
 \draw[pattern=vertical,pattern color=gray,hatch distance=8pt, hatch thickness = 0.8pt](1.5,0) rectangle (3,1.5)
node[midway]{$90^\circ$};
\draw[pattern=northeast,pattern color=gray,hatch distance=10pt, hatch thickness = 0.8pt]  (0,0) rectangle (1.5,1.5)
node[midway]{$135^\circ$};

\draw (1.5,3.5) node[below]{Base Image};

\begin{scope}[shift={(4,0)}]

\draw[pattern=horizontal,pattern color=gray,hatch distance=8pt, hatch thickness = 0.8pt]  (1.5,1.5) rectangle (3,3)node[midway]{$0^\circ$}; 
 \draw[pattern=northwest,pattern color=gray,hatch distance=10pt, hatch thickness = 0.8pt] (1.5,0) rectangle (3,1.5) node[midway]{$45^\circ$};
 \draw[pattern=vertical,pattern color=gray,hatch distance=8pt, hatch thickness = 0.8pt](0,0) rectangle (1.5,1.5)
node[midway]{$90^\circ$};
\draw[pattern=northeast,pattern color=gray,hatch distance=10pt, hatch thickness = 0.8pt]  (0,1.5) rectangle (1.5,3)
node[midway]{$135^\circ$};

\draw (1.5,3.5) node[below]{Transformed ($+90^\circ$)};

\end{scope}

\begin{scope}[shift={(8,0)}]

\draw[pattern=horizontal,pattern color=gray,hatch distance=8pt, hatch thickness = 0.8pt]  (0,1.5) rectangle (1.5,3)node[midway]{$0^\circ$}; 
 \draw[pattern=northwest,pattern color=gray,hatch distance=10pt, hatch thickness = 0.8pt] (1.5,1.5) rectangle (3,3) node[midway]{$45^\circ$};
 \draw[pattern=vertical,pattern color=gray,hatch distance=8pt, hatch thickness = 0.8pt](1.5,0) rectangle (3,1.5)
node[midway]{$90^\circ$};
\draw[pattern=northeast,pattern color=gray,hatch distance=10pt, hatch thickness = 0.8pt]  (0,0) rectangle (1.5,1.5)
node[midway]{$135^\circ$};

\draw (1.5,3.5) node[below]{Regularized Image};

\end{scope}

\end{tikzpicture}
}
\end{center}

\caption{Top row: rotation on real image and impact. From left to right, the initial image, then the $+90^\circ$ rotated image and then the regularized image using the equations \ref{eq:Rot} and \ref{eq:mod}. Bottom row: illustration of the rotation procedure on a DoFP polarimetric sensor. From left to right, the sensor's pixel subset, then subset that has undergone physical rotation, and finally the regularized pixel subset using  \mbox{equations \ref{eq:Rot} and \ref{eq:mod}.} }\label{rotreal}
\end{figure}
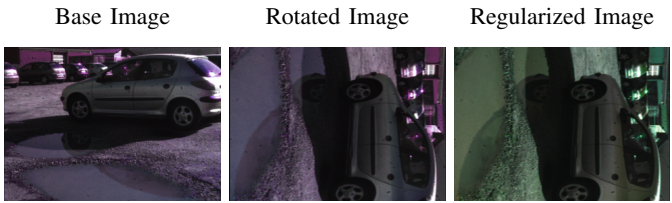
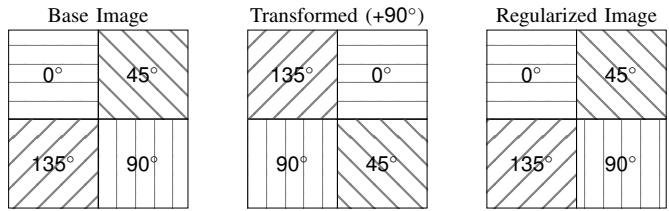
Here, we use a camera with a division of focal plane, which makes it easier to explain the influence of rotation on images. Indeed, the pixel organization due to the sensor allows preserving the integrity of the transformation by validating their order. As shown in  bottom row of Fig. \ref{rotreal}, the images are composed of super-pixels of size $2\times2$. Therefore, an isolated rotation without transformation is impossible since this would mix the order of the polarizer angles.
To however apply a rotation to the image, we need an additional process to "realign" the pixels and keep the physical properties of the polarization angle.\\

In this case, we apply an anti-clockwise rotation to the image while at the same time applying a clockwise rotation to the polarization angle, which brings back both the physics of the scene and the sensor. Let $\theta$ be the rotation angle applied to the camera, $R_\theta$ the rotation operation and $H$ the hue channel of the image (which as a reminder corresponds to the AoP): 
\begin{equation}\label{eq:Rot}
    H_{\textrm{rotated}} = R_{\theta} (H_{prev} - 2 * \textrm{\Big{.\textbf{1}}}\theta).
\end{equation}
As shown in the Figure \ref{rotreal}, bottom row, we start from the sensor and more particularly from a set of 4 DoFP pixels. If we apply pure rotation, then the sensor will be disoriented and therefore the physical direction of each image will not be respected. The regularization applied by the equation \ref{eq:Rot} then allows the image to be redirected and thus to keep the correct alignment of each polarizer.

\begin{figure}[t]
\vspace{0.3cm}
\centering
        \begin{subfigure}[b]{0.3\linewidth}  
            \centering 
            \caption{Acquisition 0$^\circ$}
            \includegraphics[width=\linewidth]{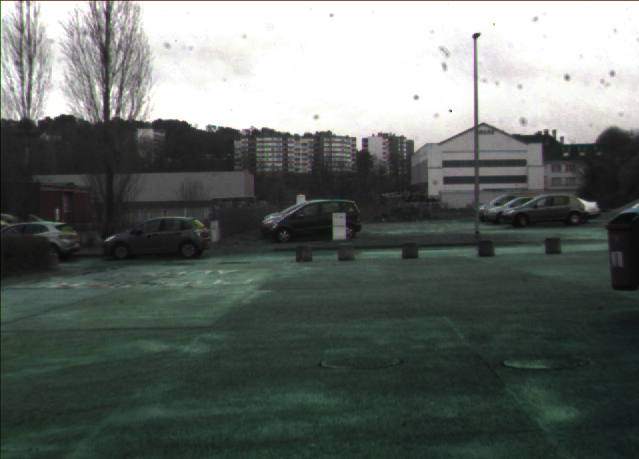}
        \end{subfigure}
        \begin{subfigure}[b]{0.3\linewidth}   
            \centering 
            \caption{Acquisition 90$^\circ$}
            \includegraphics[width=\linewidth]{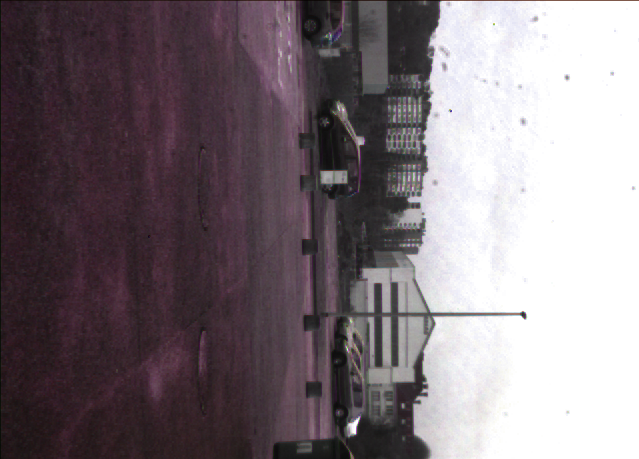}
 \end{subfigure}
        \begin{subfigure}[b]{0.3\linewidth}   
            \centering 
            \caption{Augmented 90$^\circ$}
            \includegraphics[width=\linewidth]{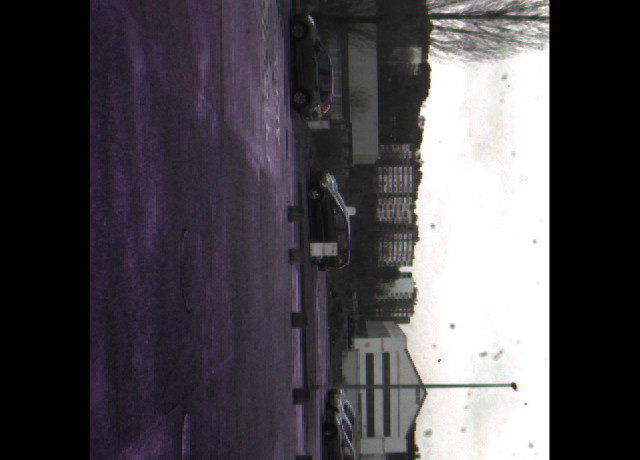}
        \end{subfigure}
        \caption{Experimental validation of augmentation process.  Images (a) and (b) are acquisitions with physical rotations applied to the camera. Image (c) is the result of augmentation applied to image (a). Note the correct recovery of angle information.
        }\label{proofexper}
        \vspace{-0.3cm}
\end{figure}

\indent The other channels are independent of the physical properties, which allows to neglect them in this transformation step. These two channels $S$ and $L$ are still subjected to the rotation operation. 
As shown in the top row of figure \ref{rotreal}, a significant modification has been made to the hue of the image when applying a 90-degree rotation augmentation.
As an experimental proof, it is equally possible to validate this procedure through different acquisitions. In the figure \ref{proofexper}, two acquisitions of the same scene were taken by applying a physical rotation to the sensor. Consequently, we obtain two images with a real rotation of 0 and 90 degrees respectively. Following the procedure described above, an artificial rotation is possible by using the 0-degree corresponding image and applying a 90-degree rotation followed by regularization. Slight visible shade differences are explained by a parallax influenced by the acquisition system. In the optimal case, the acquisition should have been conducted with exact rotation around the optical axis. Nonetheless, the shade obtained is respected, which validates the integrity of the procedure presented. 

\begin{figure}[t!]
\centering
\vspace{0.2cm}
\begin{subfigure}[b]{0.32\linewidth}
            \centering
                        \caption*{Base Image}

            \includegraphics[width=\linewidth]{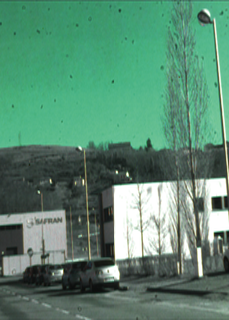}
        \end{subfigure}
        \begin{subfigure}[b]{0.32\linewidth}  
            \centering 
                        \caption*{Flipped Image}

            \scalebox{-1}[1]{\includegraphics[width=\linewidth]{images/hsl_flip.png}}
        \end{subfigure}
        \begin{subfigure}[b]{0.32\linewidth}
            \centering
                        \caption*{Regularized Image}

            \includegraphics[width=\linewidth]{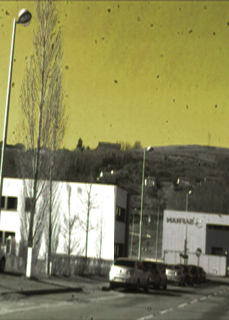}
        \end{subfigure}\\\vspace{0.2cm}\rule[1ex]{\linewidth}{0.5pt}\\\vspace{0.3cm}
        \begin{subfigure}[b]{0.29\linewidth}
            \centering
                        \caption*{\footnotesize Base Image}

            \includegraphics[width=\linewidth]{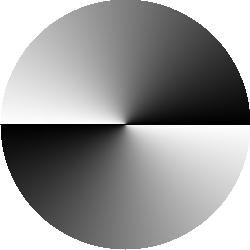}
        \end{subfigure}
        \begin{subfigure}[b]{0.29\linewidth}  
            \centering 
                        \caption*{\footnotesize Flipped Image}

            \includegraphics[width=\linewidth]{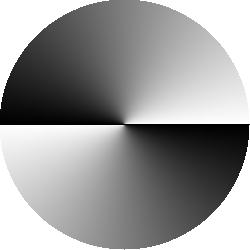}
        \end{subfigure}
        \begin{subfigure}[b]{0.29\linewidth}
            \centering
                        \caption*{\footnotesize Regularized Image}

            \includegraphics[width=\linewidth]{images/grad.jpg}
        \end{subfigure}
\caption{Top row: flipping on real image and impact. From left to right, initial image, the flipped image and the regularized image using the equations \ref{eq:flip} and \ref{eq:mod}. Bottom row: illustration of the flipping procedure. From left to right, the base image, a flipped image and the regularized image using the \mbox{equations \ref{eq:flip} and \ref{eq:mod}.}}\label{flipread}
\vspace{-0.3cm}
\end{figure}

Another possibility of augmentation proposed in this paper is symmetry. 
To ensure the integrity of the physical properties of the image, symmetry is applies as follows:
\begin{equation}\label{eq:flip}
    H_{\textrm{flipped}} = -H_{prev}.
\end{equation} 
As before for rotation, we can graphically show in Fig. \ref{flipread} the impact of pure flipping and the action of regularization.
As shown in bottom row of Fig.\ref{flipread} the right and left images seem to be the same, but unlike visual effect induced by the images particularity, the right image is a flipped version of the left image. This type of image is a grey level representation of the Hue channel. Each point of the circle in the image represents its corresponding angle with the center as reference. In conclusion, if only the top of the left circle is considered, from left to right, we have a circular gradient ranging from 180 degrees to 0 degrees. Therefore, this image represents the complete range of possible linear polarization angles in an image.
Knowing that the Hue channel of images is 360 degrees periodic, flipping consists in reversing the selected axis for the transformation.\\ The inversion of the values of the Hue channel is made possible by the use of the periodic 360 degree property by using consecutively the equations \ref{eq:flip} and \ref{eq:mod}. This procedure is illustrated in Fig. \ref{flipread}.\\

In both augmentation operations previously presented, it is necessary to maintain a consistency with respect to the properties of the color space. The H channel being a value between 0 and 360, it is then necessary to normalize the intensities: 
\begin{equation}\label{eq:mod}
H_{final} = H_{transformed} \pmod{360}
\end{equation} 
where $H_{final}$ is the final image and $H_{transformed}$ is the image resulting from a rotation, symmetry or a combination of both.

To verify the previous statements, it is possible to retrieve each polarizer intensity to verify the integrity of the information calculated as follows:
\begin{equation}
I_{\theta} = \frac{\iota}{2} (1 + \rho \cos(2\phi - 2\theta)),
\end{equation}
where $I_\theta$ is the intensity image from the polarizer oriented at angle $\theta$, $\iota$ is the intensity calculated in equation \ref{eq:I}, $\rho$ the DoP calculated in equation \ref{eq:DoPcalc} and $\phi$ the AoP deduced from equation \ref{eq:AoPcalc}.
Then the reconstruction of the images allows to examine the physical properties by calculating  the intensity obtained for each orientation of the polarizer and arranging them as on the original sensor. This procedure verifies that the addition of regularization is necessary and that without this step, the image is altered.

\section{Experiments}\label{exp}

\subsection{Implementation details}
In this section, different results from several augmentation procedures on an unique video sequence are presented. The idea here is to prove the augmentation reliability and show its contribution in improving segmentation results.\\

\begin{figure}[!tb]
\vspace{0.2cm}
\centering
 \includegraphics[width=\linewidth]{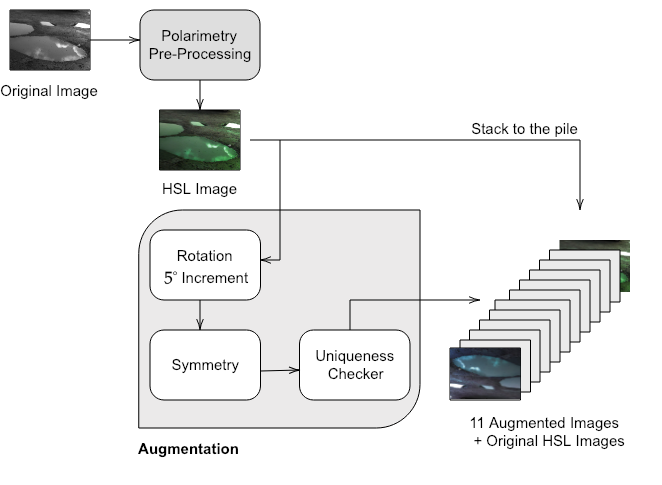}
   \caption{Illustration of the augmentation per image procedure. This process is repeated for each image in the original dataset to obtain a consistent large dataset. Then, the entire set of augmented images is shuffled.}\label{augment}
\end{figure}

\textbf{Datasets}: The training dataset is unevenly divided into a part dedicated to training and a part dedicated to validation. The validation set is composed of 50 images selected for their characteristics allowing to have a balanced representation of the classes in the images and thus to have more precise validation metrics. As for the rest of the dataset, it is employed for training and will not be balanced so that it is closer to real world data.\\
Initially, the training data set is composed of 178 images that are then either kept as they are, or augmented in a standard way, without regularization, or augmented by following our approach. Whether the dataset is increased with our method or not, the transformations involved are identical and the set accumulates a total of 2136 images. 
 As shown in Fig.\ref{augment}, the augmentation is performed by applying a random rotation in increments of 5$^\circ$ and/or symmetry with a probability of 20\%. Note that each augmentation is unique.

The test dataset is a video sequence comprised of 8,049 images acquired at a frequency of 10Hz sharing many characteristics with the training dataset. It was acquired by mounting the Trioptics PolarCam 4D Technology V polarimetric camera on a remotely operated Robotnik Summit XL robot and using the ROS operating system.\\

\textbf{Network training}: 
The experiments are conducted using a server composed of an Nvidia Titan Xp (12GB memory) GPU, 128GB of RAM and two CPU accumulating a total of 24 physical cores (48 threads).
We use DeepLabV3+\cite{chen2017rethinking} network, either with pre-trained parameters or without.
Indeed, it is equally significant to compare the influence of pre-training. Since we use DeepLabV3+, it is possible to pre-train the xception subnetwork using the most recent provided model\footnote{https://data.lip6.fr/cadene/pretrainedmodels/}.
The hyper-parameters and the loss function are kept identical for both models (pre-trained or not). We set empirical parameters like epoch number to 150, learning rate to $\eta=10^{-2}$, batch size to 8 and use Adam algorithm as optimizer.

With regard to the loss, here an adapted S\o rensen-Dice index is used to take into account the fact that some classes are under-represented in the dataset:
\begin{equation}
\lambda =\frac{ \sum^{N}_{c} 1 - \frac{2 | X_c \cap Y_c|}{|X_c| + |Y_c|}}{N},\label{eq_loss}
\end{equation}
with  $X$  the label, $Y$ the prediction, $c$ the class and $N$ the number of classes. Since classes are unequally represented in the dataset, this metric allows an equal valuation of each of them unlike other losses. \\

\begin{figure*}[!t]
\vspace{0.3cm}
\centering
                    
                      \begin{subfigure}[b]{0.18\linewidth}   
            \centering 
            \includegraphics[width=\linewidth]{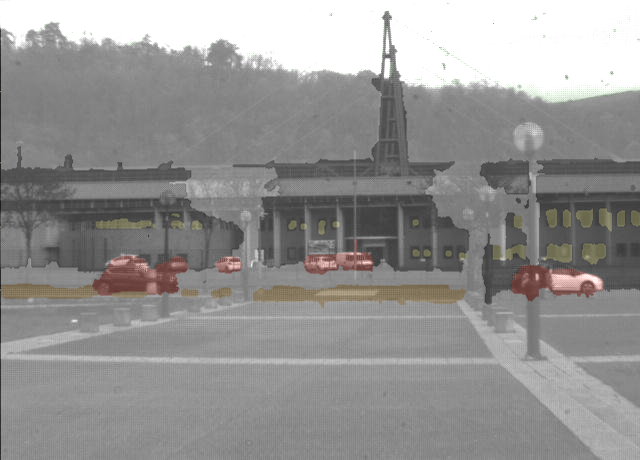}
        \end{subfigure}
        \begin{subfigure}[b]{0.18\linewidth}
            \centering
            \includegraphics[width=\linewidth]{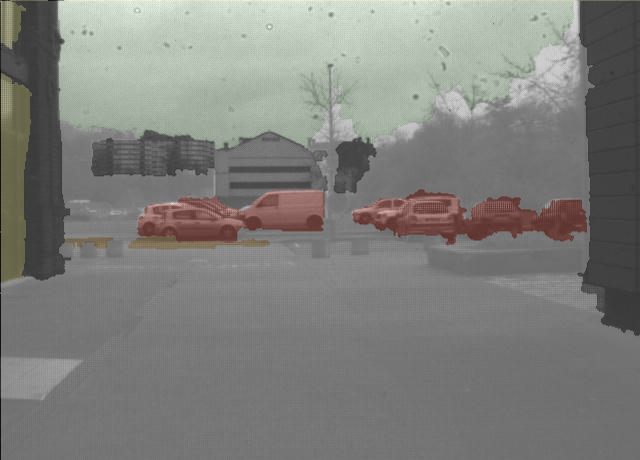}
        \end{subfigure}
        \begin{subfigure}[b]{0.18\linewidth}  
            \centering 
            \includegraphics[width=\linewidth]{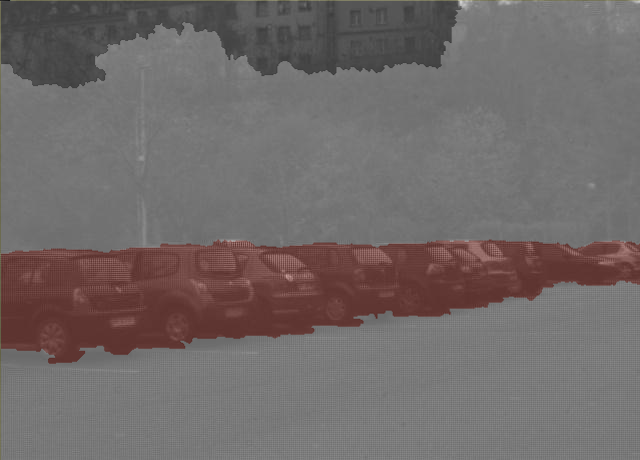}
        \end{subfigure}
        \begin{subfigure}[b]{0.18\linewidth}   
            \centering 
            \includegraphics[width=\linewidth]{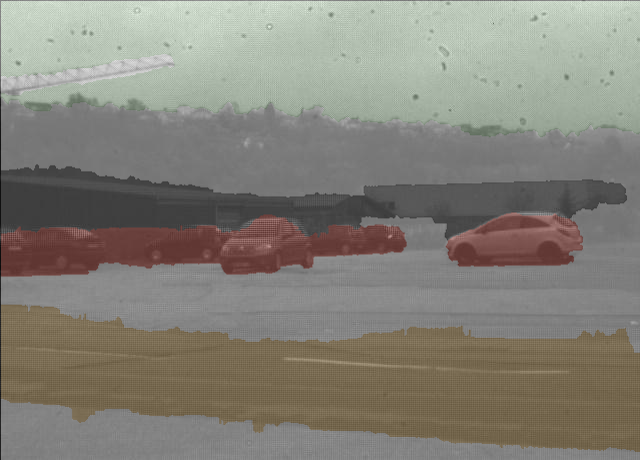}
 \end{subfigure}
        \begin{subfigure}[b]{0.18\linewidth}   
            \centering 
            \includegraphics[width=\linewidth]{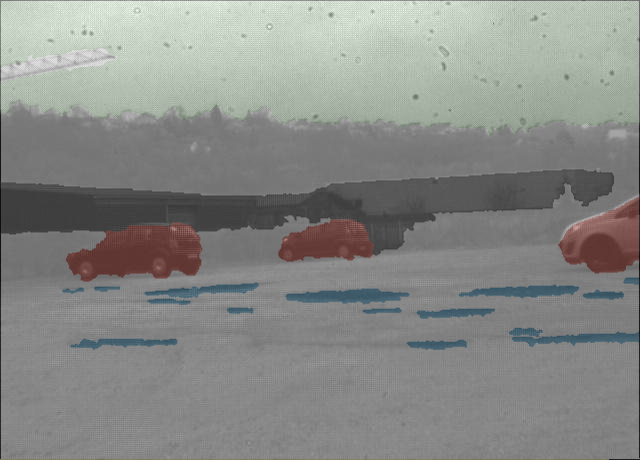}
        \end{subfigure}

        \vspace{0.1cm}

                      \begin{subfigure}[b]{0.18\linewidth}   
            \centering 
            \includegraphics[width=\linewidth]{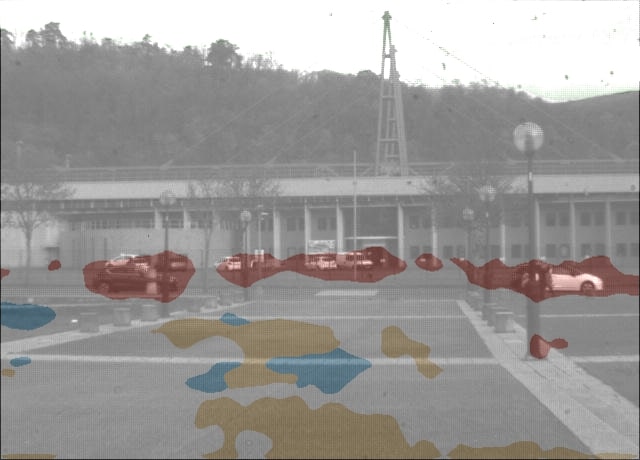}
        \end{subfigure}
        \begin{subfigure}[b]{0.18\linewidth}
            \centering
            \includegraphics[width=\linewidth]{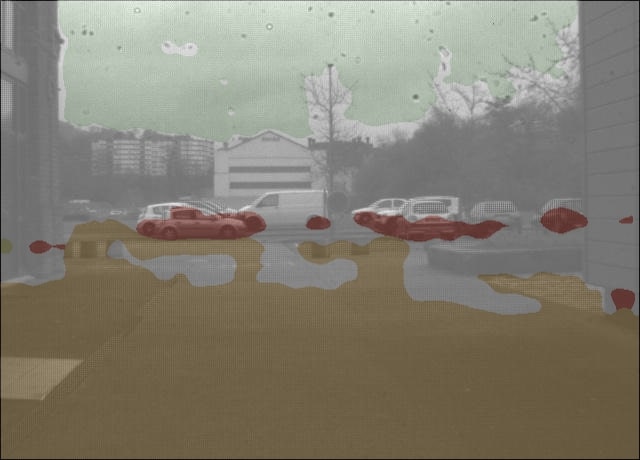}
        \end{subfigure}
        \begin{subfigure}[b]{0.18\linewidth}  
            \centering 
            \includegraphics[width=\linewidth]{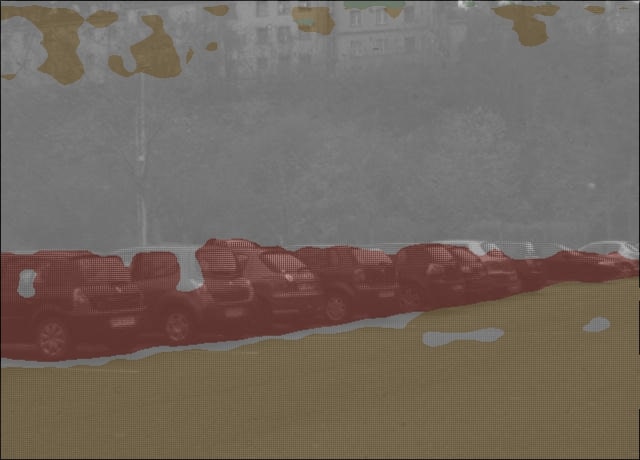}
        \end{subfigure}
        \begin{subfigure}[b]{0.18\linewidth}   
            \centering 
            \includegraphics[width=\linewidth]{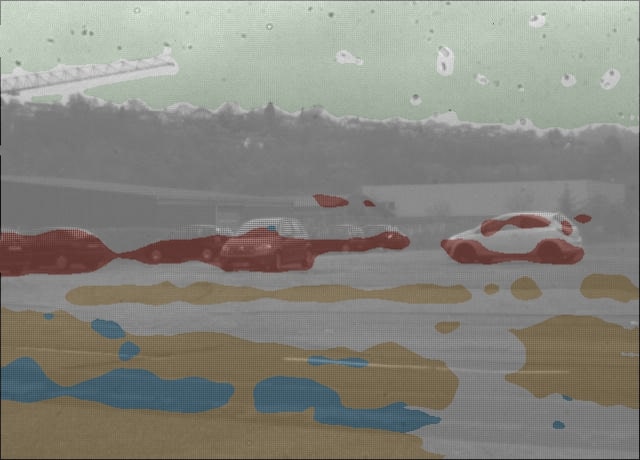}
 \end{subfigure}
        \begin{subfigure}[b]{0.18\linewidth}   
            \centering 
            \includegraphics[width=\linewidth]{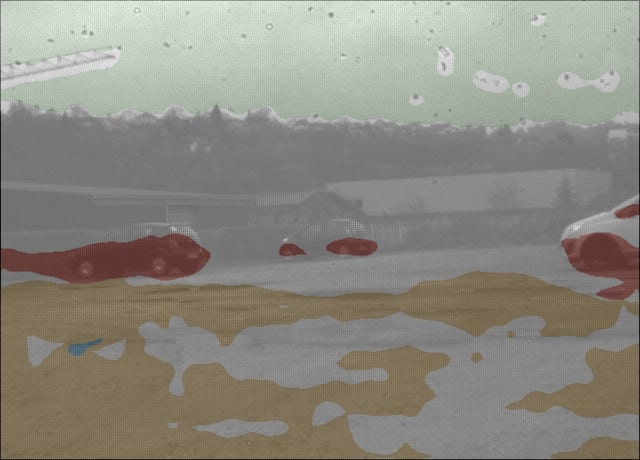}
        \end{subfigure}
     
        \vspace{0.1cm}  
        
                      \begin{subfigure}[b]{0.18\linewidth}   
            \centering 
            \includegraphics[width=\linewidth]{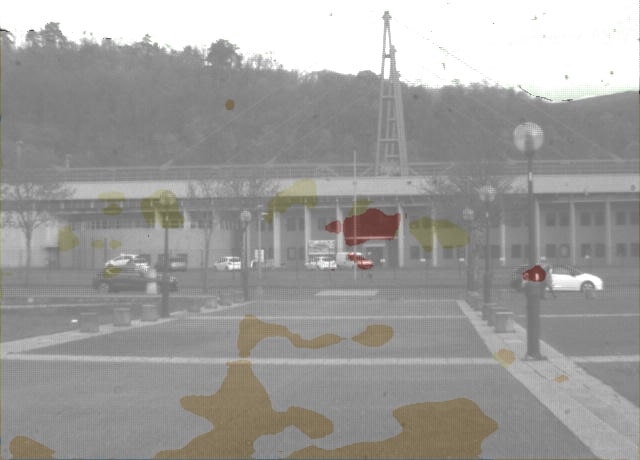}
        \end{subfigure}
        \begin{subfigure}[b]{0.18\linewidth}
            \centering
            \includegraphics[width=\linewidth]{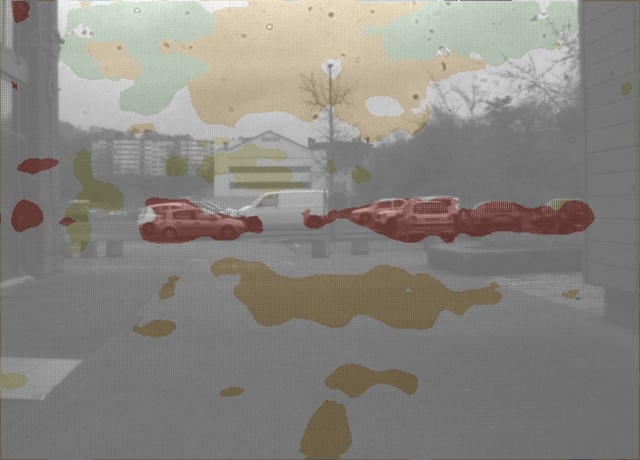}
        \end{subfigure}
        \begin{subfigure}[b]{0.18\linewidth}  
            \centering 
            \includegraphics[width=\linewidth]{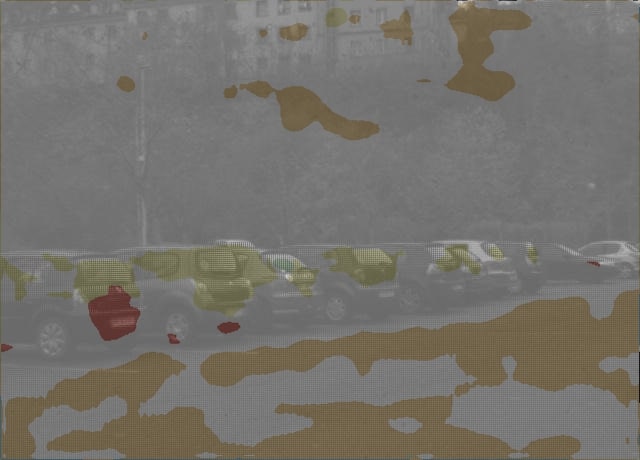}
        \end{subfigure}
        \begin{subfigure}[b]{0.18\linewidth}   
            \centering 
            \includegraphics[width=\linewidth]{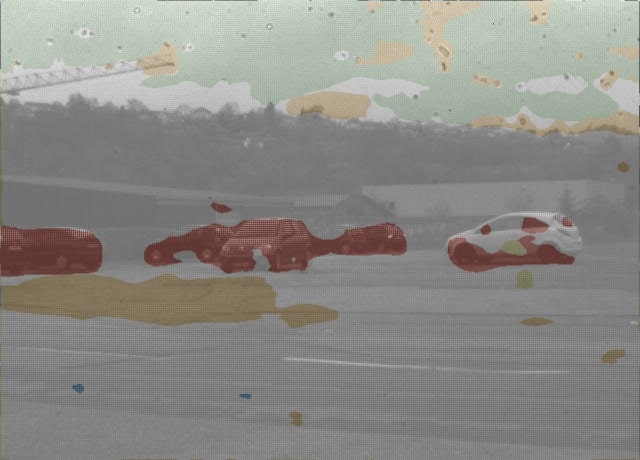}
 \end{subfigure}
        \begin{subfigure}[b]{0.18\linewidth}   
            \centering 
            \includegraphics[width=\linewidth]{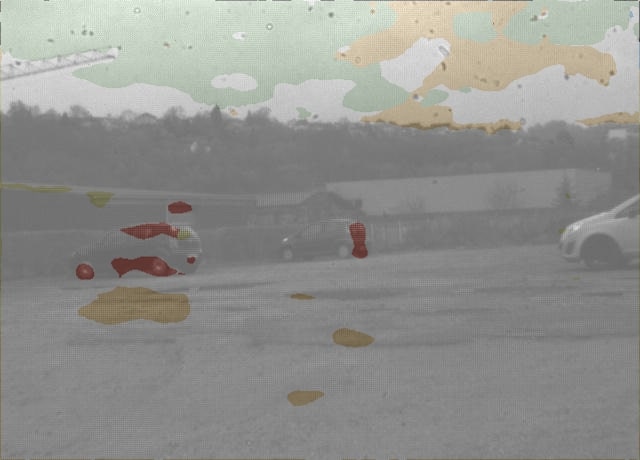}
        \end{subfigure}
 
        \vspace{0.1cm}

                        \begin{subfigure}[b]{0.18\linewidth}   
            \centering 
            \includegraphics[width=\linewidth]{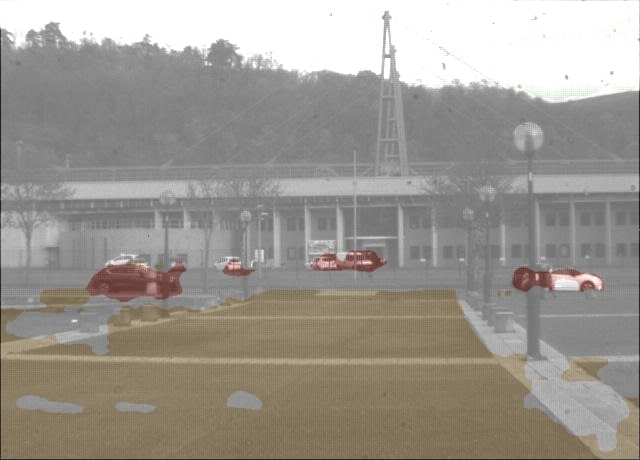}
        \end{subfigure}
        \begin{subfigure}[b]{0.18\linewidth}
            \centering
            \includegraphics[width=\linewidth]{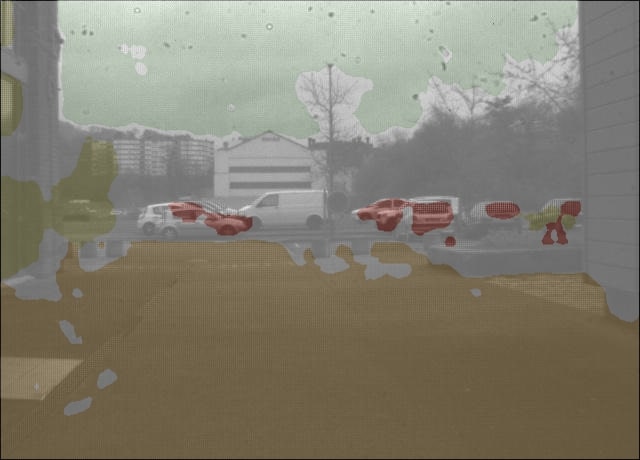}
        \end{subfigure}
        \begin{subfigure}[b]{0.18\linewidth}  
            \centering 
            \includegraphics[width=\linewidth]{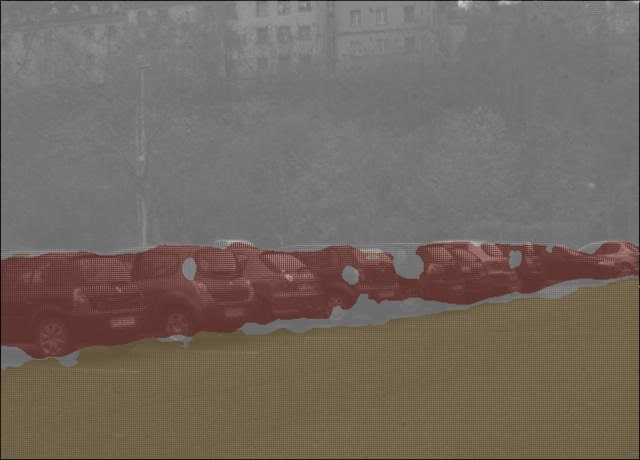}
        \end{subfigure}
        \begin{subfigure}[b]{0.18\linewidth}   
            \centering 
            \includegraphics[width=\linewidth]{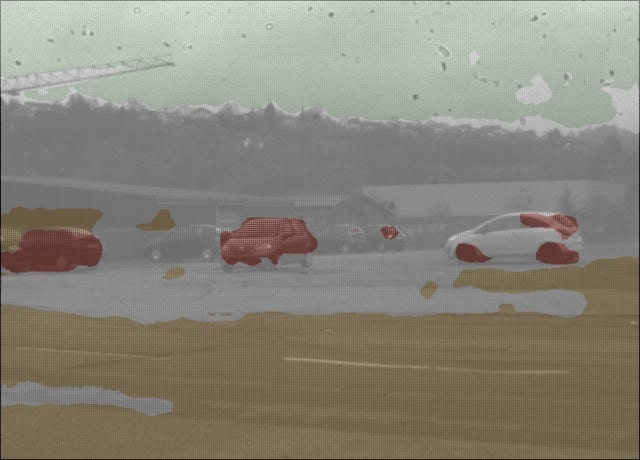}
 \end{subfigure}
        \begin{subfigure}[b]{0.18\linewidth}   
            \centering 
            \includegraphics[width=\linewidth]{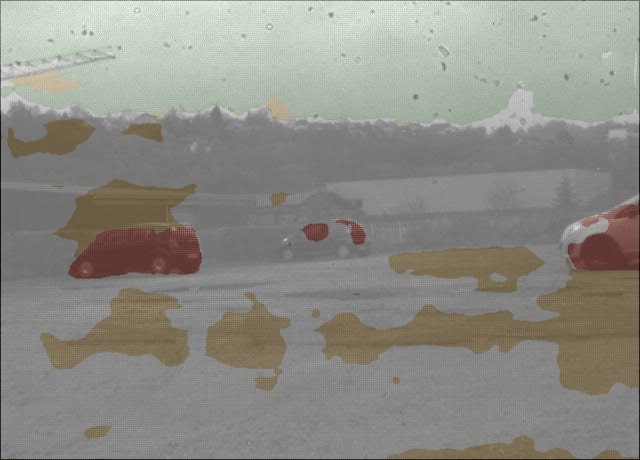}
        \end{subfigure}
        
\caption{Examples of segmentation results according to the augmentation methods. From top to bottom are present the ground-truth, then consecutively the results from model with pre-training with: no augmentation, standard augmentation and regularized augmentation.}
\label{figres}
\end{figure*}

\textbf{Metrics}: Apart from qualitative evaluation such as Fig. \ref{figres}, we have calculated a comprehensive range of metrics to quantitatively compare the different models according to augmentation processes. Results are shown in \mbox{Tab. \ref{metriques}} and correspond to metrics computed on a sub-set composed of 15 different images where all classes are represented. This avoids redundancies and therefore allows for a more detailed case-by-case examination. 
The scenes correspond to urban areas with seven different \mbox{classes: \textit{Road} (\textcolor{road}{dark yellow/orange})}, \textit{Buildings} (\textcolor{buildings}{grey}), \textit{Cars} (\textcolor{car}{red}), \textit{Water} (\textcolor{water}{blue}), \textit{Windows} (\textcolor{windows}{light yellow}), \textit{Sky} (\textcolor{sky}{green}) and \textit{None} (\textcolor{none}{light grey}).\\

\subsection{Discussion}
In this section, we will discuss the different impacts of the augmentation on the network. As shown in Fig. \ref{figres}, a brief visual assessment makes it difficult to observe substantial differences between the models even if the regularized procedure seems more appropriate. It is noteworthy that, when using the proposed augmentation method, visual aberrations are reduced. 
A predominant visual defect of all models is the absence of building detection. This will be discussed further below.\\

\renewcommand{\arraystretch}{1.5}
\begin{table*}[ht]
\centering

\caption{\label{metriques}Impact of the augmentation procedure on DeepLabV3+ network. Specific classes have been highlighted in relation to the robotic application to witness the obstacle-wise performance. Due to the limited training, \textit{Buildings} are almost undetected. For this reason, the averages denoted $\backslash B$ exclude the \textit{Buildings} class from the calculation. }
\large \center
\resizebox{\textwidth}{!}{%
\begin{tabular}{c|c|ccccc|ccccc|c|c}
\multirow{2}{*}{\textbf{Augmentation}} & \multirow{2}{*}{\textbf{PreTraining}} & \multicolumn{5}{c|}{\textbf{IoU (\%)}}                                         & \multicolumn{5}{c|}{\textbf{Recall (\%)}}                                      & \multirow{2}{*}{\textbf{Precision (\%)}} & \multirow{2}{*}{\textbf{Specificity (\%)}} \\ \cline{3-12}
                                       &                                       & \textcolor{water}{@water}        & \textcolor{windows}{@windows}      & \textcolor{car}{@cars}         & Mean          & Mean $\backslash B$ & \textcolor{water}{@water}        & \textcolor{windows}{@windows}      & \textcolor{car}{@cars}         & Mean          & Mean $\backslash B$ &                                          &                                                            \\ \hline
\multirow{2}{*}{None}                  & No                                    & 40.0          & 20.6          & 20.8          & 30.5          & 32.2           & 35.2          & 15.8          & 22.5          & \textbf{50.9} & 50.0           & 50.0                                     & 89.6                                                                           \\
                                       & Yes                                   & 54.0          & 10.3          & 43.46         & 33.5          & 34.8           & \textbf{42.4 }         & 15.3          & 57.4          & 43.3          & 50.3           & 50.1                                     & 91.0                                                                        \\ \hline
\multirow{2}{*}{Standard}              & No                                    & 0.1           & 3.4           & 12.4          & 14.8          & 13.1           & 35.0          & 25.8          & 15.0          & 31.8          & 28.0           & 41.7                                     & 88.7                                                                        \\
                                       & Yes                                   & 10.2          & 3.0           & 19.7          & 21.8          & 20.0           & 35.2          & 22.9          & 23.4          & 37.0          & 33.4           & 41.2                                     & 91.2                                                                        \\ \hline
\multirow{2}{*}{Regularized}           & No                                    & 63.9          & 13.3          & 46.7          & \textbf{43.4} & \textbf{50.3}  & 39.2 & 21.9          & \textbf{60.8} & 43.4          & \textbf{50.5}  & 48.5                                     & \textbf{91.3}                                                                \\
                                       & Yes                                   & \textbf{70.0} & \textbf{26.6} & \textbf{47.1} & 37.8          & 38.5           & 35.0          & \textbf{26.0} & 48.0          & 42.0          & 38.5           & \textbf{53.7}                            & 90.7                                                              
\end{tabular}%
}
\vspace{-0.25cm}

\end{table*}

As shown in Tab. \ref{metriques}, a large panel of metrics has been calculated to evaluate the performance of each model. The two major metrics are mean intersection over union and recall.

For each of the major metrics, we have selected three classes representing reflective areas for comparison: \textit{Water}, \textit{Windows} and \textit{Cars}. 
Class-specific metrics are, from a robotics application perspective, the critical points and also represent the core objective since they are derived from the detection of danger zones. In addition, two more general metrics are proposed to evaluate the models from various angles: precision and specificity.

From a general point of view, the Tab. \ref{metriques} points out that the regularized augmentation allows for better results in the vast majority of cases. 
In more detail, the IoU shows that models with an appropriate dataset perform better and mainly in class-specific metrics where we can observe substantial differences. On the contrary, the standard augmentation produces the worst results in terms of IoU. Metrics also emphasize it is better to use polarimetric data without augmentation than to augment them by neglecting physical properties of the scene. Indeed, with an unadapted augmentation method, the mean IoU is about twice lower than the result without augmentation. With the proposed augmentation approach, the mean IoU is improved by more than 40\%.

The recall considers the model's ability to correctly classify zones independently of bad assignments. In our case, the recall ratio per class will indicate the ability to perceive hazardous areas in general. Once more, the majority of the high results are obtained by the model with regularized data while the unsatisfactory results are held by the trained model with standard augmentation. However, for the class \textit{Water}, we observe that the model without augmentation achieves a better recall. This can be explained by the tendency for this model to overuse this class (this phenomenon is visible Fig. \ref{figres}, row 2 columns 1 and 4).


As stated above, the common defect of the six models compared is the inability to detect buildings. This difficulty can be explained by the "physical similarity" of the components forming the \textit{None} and \textit{Buildings} classes. As the training was limited, it is likely that more training epochs would have benefited for this specific class. In our approach to detection for robotics, this defect is uncritical since the class is not drastically classified as a danger. Plus, it can be deduced by exploiting the other correctly segmented classes allowing the implementation of a system of rules or constraints.\\

We can conclude the discussion by arguing that polarimetry allows for better detection of areas prone to reflection. This capability could benefit robotic applications by improving existing algorithms.
Indeed, since reflections are characterized upstream of the network, the new learnt features are specific to the reflectivity of the surfaces in the scene.
This hypothesis is validated by the unsatisfactory results obtained while neglecting physical properties of the modality and moreover when analyzing the segmentation of reflective areas like cars and waters.

\section{Conclusion}\label{sec:conclusion}
Highly reflective areas induce errors while navigating using RGB sensors. To overcome this issue, one solution is to rely on polarimetric images; however, this solution also suffers from the lack of images to train deep models and we therefore develop a successful data augmentation technique that take into account the vector aspect of polarimetric images. We proposed several regularization processes that maintain the integrity of the physical properties involved in polarization imaging. 
The experimental results show that the process of augmentation is useful and required when data amount is insufficient, but also that the results obtained at the end of the pipeline are improved in a significant way.
To go further and increase the algorithm's capabilities, other transformations such as distortion and noise addition can be investigated.
Nonetheless, the use of a new modality could be beneficial for robotics and autonomous vehicles. It has been demonstrated that, with polarimetric modality, important areas for these domains can be segmented more effectively if physical properties are unaltered. 
Moreover, it is also possible to reconsider the structure of the CNNs used to integrate the pre-processing step in order to avoid the HSL mapping of the polarized components.
\section*{Acknowledgments}

This work was supported by the French National Research Agency through ANR ICUB (ANR-17-CE22-0011). We gratefully acknowledge the support of NVIDIA Corporation with the donation of GPUs used for this research.

\bibliographystyle{IEEEtran}
\bibliography{refs}

\end{document}